\documentclass[11pt,a4paper]{article}
\PassOptionsToPackage{hyphens}{url}\usepackage{hyperref}
\usepackage[hyperref]{acl2019}
\usepackage{times}
\usepackage{latexsym}
\usepackage{graphicx}
\usepackage{subcaption}
\usepackage{amsmath}

\aclfinalcopy 

\title{Simple and Effective Text Matching with Richer Alignment Features}

\author{
  Runqi Yang$^1$, 
  Jianhai Zhang$^2$, Xing Gao$^2$, 
  Feng Ji$^2$, Haiqing Chen$^2$  \\
  $^1$Department of Computer Science and Technology, Nanjing University, China\\
  \texttt{runqiyang@gmail.com}\\
  $^2$Alibaba Group, Hangzhou, China\\
  \texttt{\{tanfan.zjh,gaoxing.gx,zhongxiu.jf,}\\
  \texttt{haiqing.chenhq\}@alibaba-inc.com}}
\date{}

\begin{document}
\maketitle
\begin{abstract}
  In this paper, we present a fast and strong neural approach for general purpose text matching applications. 
  We explore what is sufficient to build a fast and well-performed text matching model and propose to keep three key features available for inter-sequence alignment: original point-wise features, previous aligned features, and contextual features while simplifying all the remaining components.
  We conduct experiments on four well-studied benchmark datasets across tasks of natural language inference, paraphrase identification and answer selection. The performance of our model is on par with the state-of-the-art on all datasets with much fewer parameters and the inference speed is at least 6 times faster compared with similarly performed ones.
\end{abstract}

\section{Introduction}

Text matching is a core research area in natural language processing with a long history.
In text matching tasks, a model takes two text sequences as input and predicts a category or a scala value indicating their relationship.
A wide range of tasks, including natural language inference (also known as recognizing textual entailment) \cite{snli, scitail}, paraphrase identification \cite{wang2017bilateral}, answer selection \cite{wikiqa}, and so on, can be seen as specific forms of text matching problems. 
Research on general purpose text matching algorithm is beneficial to a large number of relevant applications.

Deep neural networks are the most popular choices for text matching nowadays. 
Semantic alignment and comparison of two text sequences are the keys in neural text matching. 
Many previous deep neural networks contain a single inter-sequence alignment layer. 
To make full use of this only alignment process, the model has to 
take rich external syntactic features or hand-designed alignment features as additional inputs of the alignment layer \cite{chen2017enhanced, gong2018natural},
adopt a complicated alignment mechanism \cite{wang2017bilateral, tan2018multiway}, 
or build a vast amount of post-processing layers to analyze the alignment result \cite{tay2018compare, gong2018natural}.

More powerful models can be built with multiple inter-sequence alignment layers.
Instead of making a prediction based on the comparison result of a single alignment process, a stacked model with multiple alignment layers maintains its intermediate states and gradually refines its predictions. 
However, suffering from inefficient propagation of lower-level features and vanishing gradients, these deeper architectures are harder to train. Recent works have come up with ways of connecting stacked building blocks including dense connection \cite{tay2018co, kim2018semantic} and recurrent neural networks \cite{liu2018stochastic}, which strengthen the propagation of lower-level features and yield better results than those with a single alignment process.

This paper presents RE2, a fast and strong neural architecture with multiple alignment processes for general purpose text matching. 
We question the necessity of many slow components in text matching approaches presented in previous literature, including complicated multi-way alignment mechanisms, heavy distillations of alignment results, external syntactic features, or dense connections to connect stacked blocks when the model is going deep. 
These design choices slow down the model by a large amount and can be replaced by much more lightweight and equally effective ones. 
Meanwhile, we highlight three key components for an efficient text matching model. These components, which the name RE2 stands for, are
previous aligned features (\textbf{R}esidual vectors), original point-wise features (\textbf{E}mbedding vectors), and contextual features (\textbf{E}ncoded vectors). The remaining components can be as simple as possible to keep the model fast while still yielding strong performance.

The general architecture of RE2 is illustrated in Figure \ref{fig:model}. An embedding layer first embeds discrete tokens. Several same-structured blocks consisting of encoding, alignment and fusion layers then process the sequences consecutively. These blocks are connected by an augmented version of residual connections (see section \ref{sec:a-res}). A pooling layer aggregates sequential representations into vectors which are finally processed by a prediction layer to give the final prediction. The implementation of each layer is kept as simple as possible, and the whole model, as a well-organized combination, is quite powerful and lightweight at the same time.

Our proposed method achieves the performance on par with the state-of-the-art on four benchmark datasets across three different tasks, namely SNLI and SciTail for natural language inference, Quora Question Pairs for paraphrase identification, and WikiQA for answer selection. Furthermore, our model has the least number of parameters and the fastest inference speed in all similarly-performed models. We also conduct an ablation study to compare with alternative implementations of most components, perform robustness checks to see whether the model is robust to changes of structural hyperparameters, explore what roles the three key features in RE2 play by comparing their occlusion sensitivity and show the evolution of alignment results by a case study. We release the source code\footnote{\url{https://github.com/hitvoice/RE2}, under the Apache License 2.0.} of our experiments for reproducibility and hope to facilitate future researches.

\section{Our Approach}

\begin{figure}
  \includegraphics[width=\linewidth]{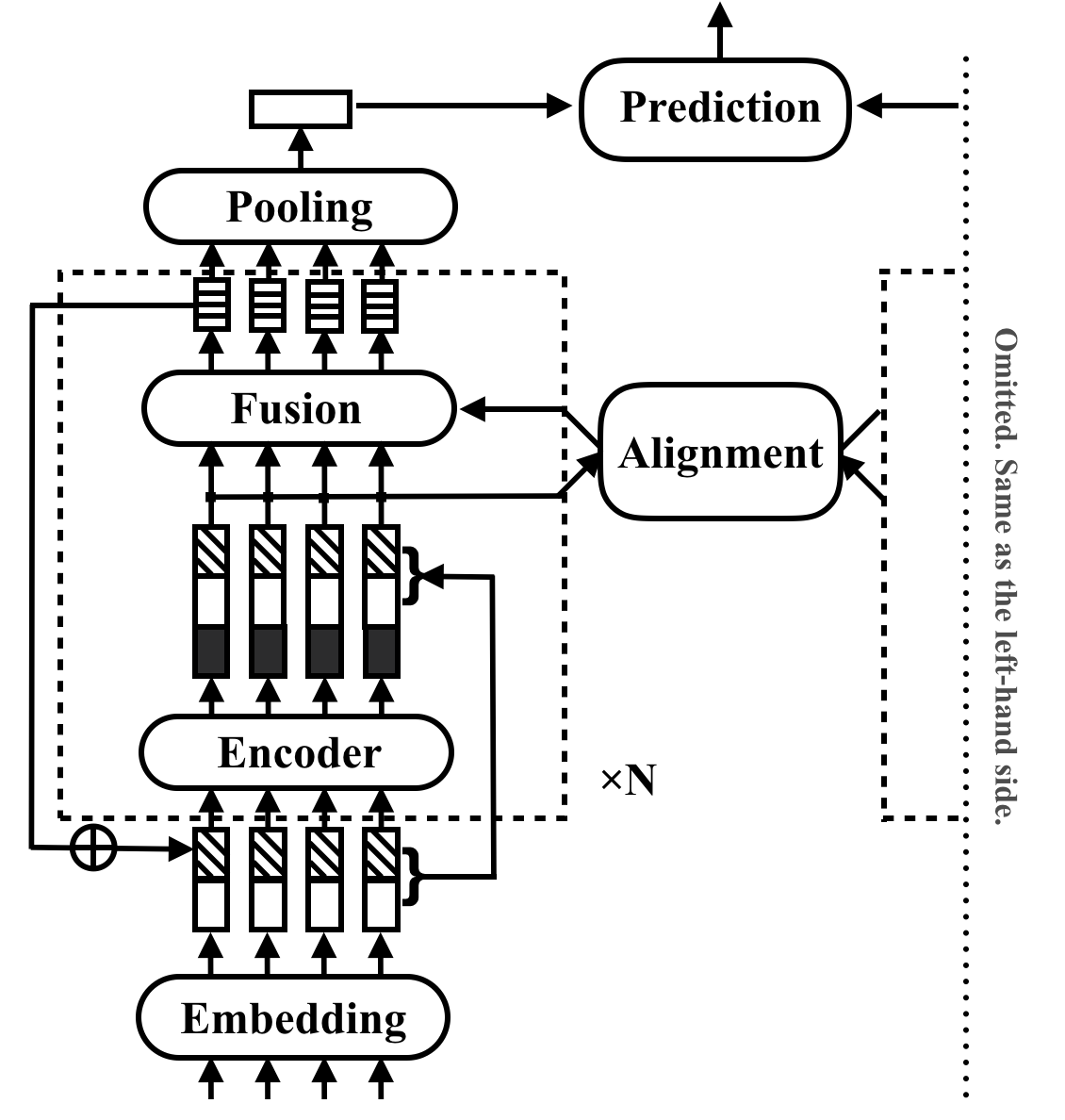}  
  \caption{An overview of RE2. There are three parts in the input of alignment and fusion layers: original point-wise features (\textbf{E}mbedding vectors, denoted by blank rectangles), previous aligned features (\textbf{R}esidual vectors, denoted by rectangles with diagonal stripes), and contextual features (\textbf{E}ncoded vectors, denoted by solid rectangles). The architecture on the right is the same as the one on the left so it's omitted for conciseness.}
  \label{fig:model}
\end{figure}

In this section, we introduce our proposed approach RE2 for text matching. Figure \ref{fig:model} gives an illustration of the overall architecture. 
Two text sequences are processed symmetrically before the prediction layer, and all parameters except those in the prediction layer are shared between the two sequences. For conciseness, we omit the part for the other sequence in the figure.

In RE2, tokens in each sequence are first embedded by the embedding layer and then processed consecutively by $N$ same-structured blocks with independent parameters (dashed boxes in Figure \ref{fig:model}) connected by augmented residual connections. Inside each block, a sequence encoder first computes contextual features of the sequence (solid rectangles in Figure \ref{fig:model}). The input and output of the encoder are concatenated and then fed into an alignment layer to model the alignment and interaction between the two sequences. A fusion layer fuses the input and output of the alignment layer. The output of the fusion layer is considered as the output of this block. The output of the last block is sent to the pooling layer and transformed into a fixed-length vector. The prediction layer takes the two vectors as input and predicts the final target. The cross entropy loss is optimized to train the model in classification tasks.

The implementation of each layer is kept as simple as possible.
We use only word embeddings in the embedding layer, without character embeddings or syntactic features. Vanilla multi-layer convolutional networks with same padding \cite{collobert2011natural} are adopted as the encoder. Recurrent networks are slower and do not lead to further improvements, so they are not adopted here. A max-over-time pooling operation \cite{collobert2011natural} is used in the pooling layer. The details of augmented residual connections and other layers are introduced as follows.

\subsection{Augmented Residual Connections} \label{sec:a-res}
To provide richer features for alignment processes, RE2 adopts an augmented version of residual connections to connect consecutive blocks. 
For a sequence of length $l$, We denote the input and output of the $n$-th block as $x^{(n)} = (x_1^{(n)}, x_2^{(n)}, \dots, x_l^{(n)})$ and $o^{(n)} = (o_1^{(n)}, o_2^{(n)}, \dots, o_l^{(n)})$, respectively. Let $o^{(0)}$ be a sequence of zero vectors.
The input of the first block $x^{(1)}$, as mentioned before, is the output of the embedding layer (denoted by blank rectangles in Figure \ref{fig:model}). 
The input of the $n$-th block $x^{(n)}$ ($n\geq 2$), is the concatenation of the input of the first block $x^{(1)}$ and the summation of the output of previous two blocks (denoted by rectangles with diagonal stripes in Figure \ref{fig:model}):
\begin{equation}
x^{(n)}_i = [x^{(1)}_i;o^{(n-1)}_i + o^{(n-2)}_i],
\end{equation}

where $[ ; ]$ denotes the concatenation operation.

With augmented residual connections, there are three parts in the input of alignment and fusion layers, namely original point-wise features kept untouched along the way (\textbf{E}mbedding vectors), previous aligned features processed and refined by previous blocks (\textbf{R}esidual vectors), and contextual features from the encoder layer (\textbf{E}ncoded vectors). Each of these three parts plays a complementing role in the text matching process. 

\subsection{Alignment Layer}

A simple form of alignment based on the attention mechanism is used following \citeauthor{parikh2016decomposable} \shortcite{parikh2016decomposable} with minor modifications. The alignment layer, as shown in Figure \ref{fig:model}, takes features from the two sequences as input and computes the aligned representations as output. Input from the first sequence of length $l_a$ is denoted as $a=(a_1, a_2, \dots, a_{l_a})$ and input from the second sequence of length $l_b$ is denoted as $b=(b_1, b_2, \dots, b_{l_b})$.
The similarity score $e_{ij}$ between $a_i$ and $b_j$ is computed as the dot product of the projected vectors:
\begin{equation}
e_{ij}=F(a_i)^T F(b_j). \label{eq:align}
\end{equation}
$F$ is an identity function or a single-layer feed-forward network. The choice is treated as a hyper-parameter.

The output vectors $a'$ and $b'$ are computed by weighted summation of representations of the other sequence. The summation is weighted by similarity scores between the current position and the corresponding positions in the other sequence:
\begin{align}
\begin{split}
  a_i' &= \sum_{j=1}^{l_b}\frac{\exp(e_{ij})}{\sum_{k=1}^{l_b}\exp(e_{ik})}b_j, \\
  b_j' &= \sum_{i=1}^{l_a}\frac{\exp(e_{ij})}{\sum_{k=1}^{l_a}\exp(e_{kj})}a_i.
\end{split} \label{eq:aligned}
\end{align}

\subsection{Fusion Layer}
The fusion layer compares local and aligned representations in three perspectives and then fuse them together. The output of the fusion layer for the first sequence $\bar a$ is computed by
\begin{align}
  \begin{split}
  \bar a^1_i &= G_1([a_i; a_i']), \\
  \bar a^2_i &= G_2([a_i; a_i - a_i']), \\
  \bar a^3_i &= G_3([a_i; a_i \circ a_i']), \\
  \bar a_i &= G([\bar a^1_i; \bar a^2_i; \bar a^3_i]),
  \end{split}
\end{align}
where $G_1$, $G_2$, $G_3$, and $G$ are single-layer feed-forward networks with independent parameters and $\circ$ denotes element-wise multiplication. The subtraction operator highlights the difference between the two vectors while the multiplication highlights similarity. Formulations for $\bar b$ are similar and omitted here.

\subsection{Prediction Layer}
The prediction layer takes the vector representations of the two sequences $v_1$ and $v_2$ from the pooling layers as input and predicts the final target following \citeauthor{mou2016natural} \shortcite{mou2016natural}:
\begin{equation}
\hat {\mathbf y} = 
H([v_1; v_2; v_1 - v_2; v_1 \circ v_2]). 
\label{eq:standard_pred}
\end{equation}
$H$ is a multi-layer feed-forward neural network. In a classification task, $\hat {\mathbf y} \in \mathcal{R}^C$ represents the unnormalized predicted scores for all classes where $C$ is the number of classes. The predicted class is $\hat y=\operatorname{argmax}_i \hat {\mathbf y}_i$. In a regression task, $\hat {\mathbf y}$ is the predicted scala value.

In symmetric tasks like paraphrase identification, a symmetric version of the prediction layer is used for better generalization:
\begin{equation}
\hat {\mathbf y} = H([v_1; v_2; |v_1 - v_2|; v_1 \circ v_2]). 
\label{eq:symmetric_pred}
\end{equation}

We also provide a simplified version of the prediction layer. Which version to use is treated as a hyperparameter. The simplified prediction layer can be expressed as:

\begin{equation}
  \hat {\mathbf y} = H([v_1; v_2]). 
  \label{eq:simplified_pred}
\end{equation}

\section{Experiments}
\subsection{Datasets}
In this section, we briefly introduce datasets used in the experiments and their evaluation metrics.

{\bf SNLI} \cite{snli} (Stanford Natural Language Inference) is a benchmark dataset for natural language inference. In natural language inference tasks, the two input sentences are asymmetrical. The first one is called ``premise'' and the second is called ``hypothesis''. The dataset contains 570k human annotated sentence pairs from an image captioning corpus, with labels ``entailment'', ``neutral'', ``contradiction'' and ``-''. The ``-'' label indicates that the annotators cannot reach an agreement, so we ignore text pairs with this kind of labels in training and testing following \citeauthor{snli} \shortcite{snli}. We use the same dataset split as in the original paper. Accuracy is used as the evaluation metric for this dataset.

{\bf SciTail} \cite{scitail} (Science Entailment) is an entailment classification dataset constructed from science questions and answers. Since scientific facts cannot contradict with each other, this dataset contains only two types of labels, entailment and neutral. We use the original dataset partition. This dataset contains 27k examples in total. 10k examples are with entailment labels and the remaining 17k are labeled as neutral. Accuracy is used as the evaluation metric for this dataset.

{\bf Quora Question Pairs}\footnote{\url{https://data.quora.com/First-Quora-Dataset-Release-Question-Pairs}} 
is a dataset for paraphrase identification with two classes indicating whether one question is a paraphrase of the other. The dataset contains more than 400k real question pairs collected from Quora.com. We use the same dataset partition as mentioned in \citeauthor{wang2017bilateral} \shortcite{wang2017bilateral}. Accuracy is used as the evaluation metric for this dataset.

{\bf WikiQA} \cite{wikiqa} is a retrieval-based question answering dataset based on Wikipedia. It contains questions and their candidate answers, with binary labels indicating whether a candidate sentence is a correct answer to the question it belongs to. This dataset has 20.4k training pairs, 2.7k development pairs, and 6.2k testing pairs. Mean average precision (MAP) and mean reciprocal rank (MRR) are used as the evaluation metrics for this task.

\subsection{Implementation Details}
We implement our model with TensorFlow \cite{abadi2016tensorflow} and train on Nvidia P100 GPUs. We tokenize sentences with the NLTK toolkit \cite{bird2009natural}, convert them to lower cases and remove all punctuations. We do not limit the maximum sequence length, and all sequences in a batch are padded to the batch-wise maximum.
Word embeddings are initialized with 840B-300d GloVe word vectors \cite{pennington2014glove} and fixed during training. Embeddings of out-of-vocabulary words are initialized to zeros and fixed as well. All other parameters are initialized with He initialization \cite{he2015delving} and normalized by weight normalization \cite{salimans2016weight}. 
Dropout with a keep probability of 0.8 is applied before every fully-connected or convolutional layer. The kernel size of the convolutional encoder is set to 3. The prediction layer is a two-layer feed-forward network. The hidden size is set to 150 in all experiments. Activations in all feed-forward networks are GeLU activations \cite{hendrycks2016bridging}, and we use $\sqrt{2}$ as an approximation of the variance balancing parameter for GeLU activations in He initialization. We scale the summation in augmented residual connections by $1/\sqrt{2}$ when $n \geq 3$ to preserve the variance under the assumption that the two addends have the same variance.

The number of blocks is tuned in a range from 1 to 3. The number of layers of the convolutional encoder is tuned from 1 to 3. Although in robustness checks (Table \ref{tab:robustness}) we validate with up to 5 blocks and layers, in all other experiments we deliberately limit the maximum number of blocks and number of layers to 3 to control the size of the model. We use the Adam optimizer \cite{kingma2015adam} and an exponentially decaying learning rate with a linear warmup. The initial learning rate is tuned from 0.0001 to 0.003. The batch size is tuned from 64 to 512. The threshold for gradient clipping is set to 5. For all the experiments except for the comparison of ensemble models, we report the average score and the standard deviation of 10 runs.

\subsection{Results on Natural Language Inference}

Results on SNLI dataset are listed in Table \ref{tab:snli-result}. We compare single models and ensemble models. For a fair comparison, we only compare with results obtained without external contextualized embeddings. In the ensemble experiment, we train 8 models with different random seeds and ensemble the results by a voting strategy. 

Our method obtains a result on par with the state-of-the-art among single models and a highly competitive result among ensemble models, with only a few parameters. Compared to SAN, our model reduces 20\% parameters while improves the performance by 0.3\% in accuracy, which indicates that our proposed architecture is highly efficient. 

\begin{table}
  \centering
  \small
  \begin{tabular}{|l|l|l|}
  \hline
  {\bf Model} & {\bf Params} & {\bf Acc.(\%)}\\\hline
    DecAtt \cite{parikh2016decomposable} & 0.6M & 86.8 \\
    BiMPM \cite{wang2017bilateral} & 1.6M & 86.9 \\
    ESIM \cite{chen2017enhanced} & 4.3M & 88.0 \\ 
    DIIN \cite{gong2018natural} & 4.4M & 88.0 \\ 
    MwAN \cite{tan2018multiway} & 14M & 88.3 \\
    CAFE \cite{tay2018compare} & 4.7M & 88.5 \\
    HIM \cite{chen2017enhanced} & 7.7M & 88.6 \\
    SAN \cite{liu2018stochastic} & 3.5M & 88.6 \\
    CSRAN \cite{tay2018co} & 13.9M & 88.7 \\
    DRCN \cite{kim2018semantic} & 6.7M & {\bf 88.9} \\\hline
    RE2 (ours)& 2.8M & {\bf 88.9$\pm$0.1}\\\hline
    BiMPM (ensemble) & 6.4M & 88.8 \\
    DIIN (ensemble) & 17M & 88.9 \\
    CAFE (ensemble) & 17.5M & 89.3 \\
    MwAN (ensemble) & 58M & 89.4 \\
    DRCN (ensemble) & 53.3M & {\bf 90.1} \\\hline
    RE2 (ensemble) & 22.4M & 89.9 \\\hline
  \end{tabular}
  \caption{Experimental results on SNLI test set.}\label{tab:snli-result}
\end{table}

Results on Scitail dataset are listed in Table \ref{tab:scitail-result}. The performance of our method is very close to state-of-the-art. This dataset is considered much more difficult with fewer training data available and generally low accuracy as a binary classification problem. The variance of the results is larger since the size of training and test set is only 4\% and 20\% compared to those of SNLI.

\begin{table}
  \centering
  \small
  \begin{tabular}{|l|l|}
  \hline
  {\bf Model} & {\bf Acc(\%)}\\\hline
  ESIM \cite{chen2017enhanced} & 70.6 \\  
  DecompAtt \cite{parikh2016decomposable} & 72.3 \\
  DGEM \cite{scitail} & 77.3 \\
  HCRN \cite{tay2018hermitian} & 80.0 \\
  CAFE \cite{tay2018compare} & 83.3 \\
  CSRAN \cite{tay2018co} & {\bf 86.7} \\\hline
  RE2 (ours) & {\bf 86.0$\pm$0.6} \\\hline
  \end{tabular}
  \caption{Experimental results on SciTail test set.}\label{tab:scitail-result}
\end{table}

\subsection{Results on Paraphrase Identification}
Results on Quora dataset are listed in Table \ref{tab:quora-result}. Since paraphrase identification is a symmetric task where two input sequences can be swapped with no effect to the label of the text pair, in hyperparameter tuning we validate between two symmetric versions of the prediction layer (Equation \ref{eq:symmetric_pred} and Equation \ref{eq:simplified_pred}) and use no additional data augmentation. The performance of RE2 is on par with the state-of-the-art on this dataset. 

\begin{table}
  \centering
  \small
  \begin{tabular}{|l|l|}
  \hline
  {\bf Model} & {\bf Acc.(\%)}\\\hline
    BiMPM \cite{wang2017bilateral} & 88.2 \\
    pt-DecAttn-word \cite{tomar2017neural} & 87.5\\
    pt-DecAttn-char \cite{tomar2017neural} & 88.4\\
    DIIN \cite{gong2018natural} & {\bf 89.1} \\ 
    MwAN \cite{tan2018multiway} & {\bf 89.1} \\
    CSRAN \cite{tay2018co} & {\bf 89.2} \\
    SAN \cite{liu2018stochastic} & {\bf 89.4} \\\hline
    RE2 (ours) & {\bf 89.2$\pm$0.2}
    \\\hline
  \end{tabular}
  \caption{Experimental results on Quora test set. }\label{tab:quora-result}
\end{table}

\subsection{Results on Answer Selection}
Results on WikiQA dataset are listed in Table \ref{tab:wikiqa-result}. Note that some of the previous methods round their reported results to three decimal points, but we choose to align with the original paper \cite{wikiqa} and round our results to four decimal points. In hyperparameter tuning, we choose the best hyperparameters including early stopping according to MRR on WikiQA development set. We obtain a result on par with the state-of-the-art reported on this dataset. It's worth mentioning that we still train our model by point-wise binary classification loss, unlike some of the previous methods (including HCRN) which are trained by the pairwise ranking loss. Our method can perform well in the answer selection task without any task-specific modifications. 

\begin{table}
  \centering
  \small
  \begin{tabular}{|l|l|l|}
  \hline
  {\bf Model} & {\bf MAP} & {\bf MRR}\\\hline
  ABCNN \cite{yin2016abcnn} & 0.6921 & 0.7108 \\
  KVMN \cite{miller2016key} & 0.7069 & 0.7265 \\
  BiMPM \cite{wang2017bilateral} & 0.718 & 0.731 \\
  IWAN \cite{shen2017inter} & 0.733 & 0.750 \\
  CA \cite{wang2017compare} & {\bf 0.7433} & {\bf 0.7545} \\
  HCRN \cite{tay2018hermitian} & {\bf 0.743} & {\bf 0.756} \\\hline
  RE2 (ours) & {\bf 0.7452} & {\bf 0.7618}\\
  & {\bf $\pm$0.0044} & {\bf $\pm$0.0040}\\\hline
  \end{tabular}
  \caption{Experimental results on WikiQA test set.}\label{tab:wikiqa-result}
\end{table}

\subsection{Inference Time}

To show the efficiency of our proposed model, we compare the inference time with some other models whose code is open-source. Table \ref{tab:inference-time} shows the comparison results. All the compared models are implemented in TensorFlow in the original implementations. The $\dagger$ mark indicates that the model uses POS tags as external syntactic features and the computation time of POS tagging is not included. In our RE2 model, the number of encoder layers is set to 3, the largest possible number in all previously reported experiments. Besides, since all the reported results of our proposed method are obtained with no more than 3 blocks, we only measure the inference time of RE2 with 1-3 blocks. We train all the compared models using the official training code and commands released by the authors on Nvidia P100 GPUs and save model checkpoints to disk. After training, all the models are required to make predictions for a batch of 8 pairs of sentences on a MacBook Pro with Intel Core i7 CPUs. The lengths of these sentences are 20 and the maximum number of characters in a word is 12. The reported statistics are the average and the standard deviation of processing 100 batches. 

\begin{table}
  \centering
  \small
  \begin{tabular}{|l|l|}
  \hline
  {\bf Model} & {\bf time(s/batch)}\\\hline
  BiMPM \cite{wang2017bilateral} & 0.05 $\pm$ 0.00 \\
  CAFE$^\dagger$ \cite{tay2018compare} & 0.07 $\pm$ 0.01 \\
  DIIN$^\dagger$ \cite{gong2018natural} & 0.85 $\pm$ 0.11 \\
  DIIN with EM feature$^\dagger$\ & 1.79 $\pm$ 0.22 \\
  CSRAN$^\dagger$ \cite{tay2018co} & 0.28 $\pm$ 0.02 \\\hline
  RE2 (1 block) &  0.03 $\pm$ 0.00 \\
  RE2 (2 blocks) &  0.04 $\pm$ 0.00 \\
  RE2 (3 blocks) &  0.05 $\pm$ 0.00 \\\hline
  \end{tabular}
  \caption{Inference time when batch size $=8$ on Intel Core i7 CPUs. Models with $\dagger$ marks use POS tags as external syntactic features and the computation time of POS tagging is not included.}\label{tab:inference-time}
\end{table}

The comparison results in Table \ref{tab:inference-time} show that our method has very high CPU inference speed, even with multiple stacked blocks. Compared with similarly performed methods, ours is 6 times faster than CSRAN and at least 17 times faster than DIIN. With the highly efficient design, our method can perform well without any strong but slow building blocks like recurrent neural networks, dense connections or any syntactic features. Compared with models of similar inference speed, BiMPM and CAFE, ours obtains much higher prediction scores according to Table \ref{tab:snli-result}, Table \ref{tab:scitail-result}, Table \ref{tab:quora-result} and Table \ref{tab:wikiqa-result}. 

In summary, our proposed method achieves performance on par with the state-of-the-art on all four well-studied datasets across three different tasks with only a few parameters and fast inference speed. 

\subsection{Analysis}

{\bf Ablation study}. 
We present an ablation study of our model, comparing the original model with 6 ablation baselines: (1) ``w/o enc-in'': use directly the output of the encoder as the input of the alignment and fusion layers like in most previous approaches without concatenating the encoder input; (2) ``residual conn.'': use vanilla residual connections ($x^{(n)}_i = o^{(n-1)}_i + o^{(n-2)}_i$) in place of the augmented version; (3) ``simple fusion'': use simply $\bar a_i = G_1([a_i; a_i'])$ and $\bar b_i = G_1([b_i; b_i'])$ as the fusion layer; (4) ``alignment alt.'': use the alternative version of the alignment layer where $F$ in Equation \ref{eq:align} is a single-layer feed-forward network or an identity function; (5) ``prediction alt.'': use the alternative version (Equation \ref{eq:standard_pred}/\ref{eq:symmetric_pred} or Equation \ref{eq:simplified_pred}) of the prediction layer; (6) parallel blocks: feed the embeddings directly to all the blocks and sum up their outputs as the input of the pooling layer instead of processing input sequences consecutively by each block. The last setting is designed to study whether the improvement is due to deeper architecture or just a larger amount of parameters. 

The ablation study is conducted on the development set of SNLI, Quora, Scitail, and WikiQA. In WikiQA we choose MRR as the evaluation metric. Note that on SciTail, $F$ in Equation \ref{eq:align} in alignment layers is an identity function while on all other datasets $F$ is a single-layer feed-forward network. On WikiQA, the simplified version (Equation \ref{eq:simplified_pred}) is used as the prediction layer while on all other datasets the full version (Equation \ref{eq:standard_pred} or \ref{eq:symmetric_pred}) is used. The reported results are the average of 10 runs and the standard deviations are omitted for clarity.

\begin{table}
  \centering
  \small
  \begin{tabular}{|l|l|l|l|l|}
  \hline
    & {\bf SNLI} & {\bf Quora} & {\bf Scitail} & {\bf WikiQA}\\\hline
  original & 88.9 & 89.4 & 88.9 & 0.7740\\
  w/o enc-in & 87.2 & 85.7 & 78.1 & 0.7146 \\ 
  residual conn. & 88.9 & 89.2 & 87.4 & 0.7640  \\
  simple fusion & 88.8 & 88.3 & 87.5 &0.7345 \\
  alignment alt.& 88.7 & 89.3 & 88.2 & 0.7702 \\
  prediction alt.& 88.9 & 89.2 & 88.8 & 0.7558 \\
  parallel blocks & 88.8 & 88.6 & 87.6 &0.7607\\\hline
  \end{tabular}
  \caption{Ablation study on dev sets of the corresponding datasets.}\label{tab:ablation}
\end{table}

\begin{figure*}
  \centering
  \begin{subfigure}{.33\textwidth}
    \centering
    \includegraphics[width=\textwidth]{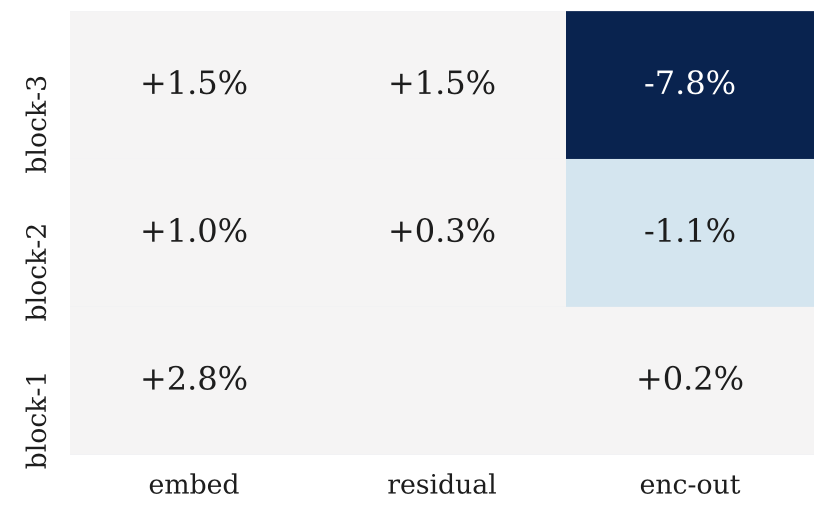}
    \caption{entailment}
    \label{fig:entailment}
  \end{subfigure}%
  \begin{subfigure}{.33\textwidth}
    \centering
    \includegraphics[width=\textwidth]{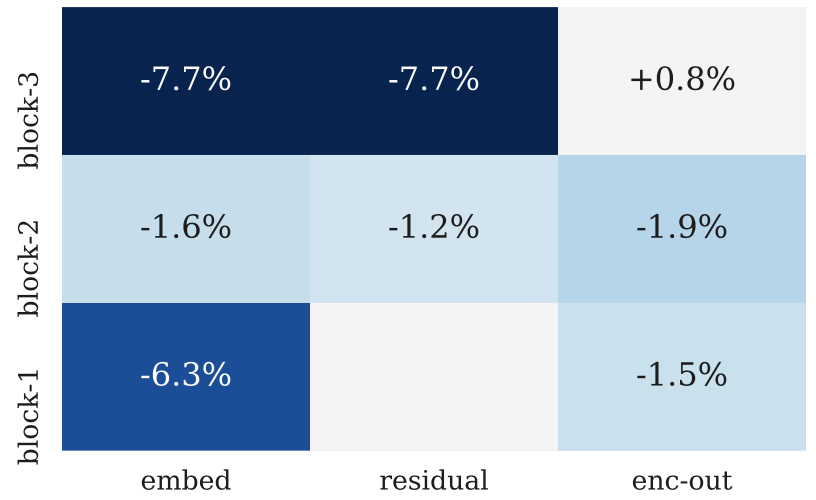}
    \caption{neutral}
    \label{fig:neutral}
  \end{subfigure}
  \begin{subfigure}{.33\textwidth}
    \centering
    \includegraphics[width=\textwidth]{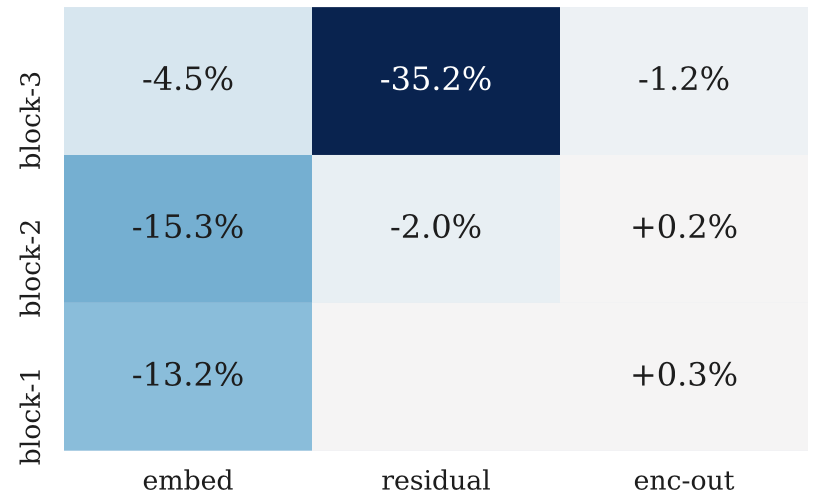}
    \caption{contradiction}
    \label{fig:contradiction}
  \end{subfigure}
  \caption{Occlusion sensitivity of different parts in the input of the alignment layers on SNLI dev set: original point-wise features (embed), aligned features (residual), and contextual features (enc-out).}
  \label{fig:occlusion}
\end{figure*}

The result is shown in Table \ref{tab:ablation}. The first ablation baseline shows that without richer features as the alignment input, the performance on all datasets degrades significantly. This is the key component in the whole model. The results of the second baseline show that vanilla residual connections without direct access to the original point-wise features are not enough to model the relations in many text matching tasks. The simpler implementation of the fusion layer leads to evidently worse performance, indicating that the fusion layer cannot be further simplified. On the other hand, the alignment layer and the prediction layer can be simplified on some of the datasets. In the last ablation study, we can see that parallel blocks perform worse than stacked blocks, which supports the preference for deeper models over wider ones.

{\bf Robustness checks}. To check whether our proposed method is robust to different variants of structural hyperparameters, we experiment with (1) the number of blocks varying from 1 to 5 with the number of encoder layers set to 2; (2) the number of encoder layers varying from 1 to 5 with the number of blocks set to 2. Robustness checks are performed on the development set of SNLI, Quora and Scitail. The result is presented in Table \ref{tab:robustness}. We can see in the table that fewer blocks or layers may not be sufficient but adding more blocks or layers than necessary hardly harms the performance. 
On WikiQA dataset, our method does not seem to be robust to structural hyperparameter changes. \citeauthor{crane2018questionable} \shortcite{crane2018questionable} mentions that on WikiQA dataset a neural matching model \cite{severyn2015learning} trained with different random seeds can result in differences up to 0.08 in MAP and MRR. We leave the further investigation of the high variance on the WikiQA dataset for further work.

\begin{table}
  \centering
  \small
  \begin{tabular}{|l|l|l|l|}
  \hline
    & {\bf SNLI} & {\bf Quora} & {\bf Scitail} \\\hline
  1 block & 88.1$\pm$0.1       & 88.7$\pm$0.1       & 88.3$\pm$0.8        \\
  2 blocks & 88.9$\pm$0.2       & 89.2$\pm$0.2       & {\bf 88.9$\pm$0.3}  \\
  3 blocks & 88.9$\pm$0.1       & 89.4$\pm$0.1       & 88.8$\pm$0.5        \\
  4 blocks & {\bf 89.0$\pm$0.1} & {\bf 89.5$\pm$0.1} & 88.7$\pm$0.5        \\
  5 blocks & 89.0$\pm$0.2       & 89.2$\pm$0.2       & 88.5$\pm$0.5        \\\hline
  1 enc. layer & 88.6$\pm$0.2       & 88.9$\pm$0.2       & 88.1$\pm$0.4       \\
  2 enc. layers & 88.9$\pm$0.2       & 89.2$\pm$0.2       & 88.9$\pm$0.3       \\
  3 enc. layers & {\bf 89.2$\pm$0.1} & {\bf 89.2$\pm$0.1} & 88.7$\pm$0.6       \\
  4 enc. layers & 89.1$\pm$0.0       & 89.1$\pm$0.1       & 88.7$\pm$0.5       \\
  5 enc. layers & 89.0$\pm$0.1       & 89.0$\pm$0.2       & {\bf 89.1$\pm$0.3} \\\hline
  \end{tabular}
  \caption{Robustness checks on dev sets of the corresponding datasets.}
  \label{tab:robustness}
\end{table}

{\bf Occlusion sensitivity}. To better understand what roles the three alignment features play, we perform an analysis of occlusion sensitivity similar to those in computer vision \cite{zeiler2014visualizing}. We use a three-block RE2 model to predict on SNLI dev set, mask one feature in one block to zeros at a time and report changes in accuracy of the three categories: entailment, neutral and contradiction. Occlusion sensitivity can help to reveal how much the model depends on each part when deciding on a specific category and we can make some speculations about how the model works based on the observations. Figure \ref{fig:occlusion} shows the result of occlusion sensitivity. Previous aligned features are absent in the first block and thus left blank.

The text matching process can be abstracted, with moderate simplifications, to three stages: aligning tokens between the two sequences, focusing on a subset of the aligned pairs, discerning the semantic relations between the attended pairs. Each of the three key features in RE2 has a closer connection with one of the stages. 

As we can see in Figure \ref{fig:entailment}, contextual features, represented by the output of the encoder, are indispensable when predicting entailment. These features connect with the first stage of text matching. The sequence encoder, implemented by convolutional networks, models local and phrase-level semantics, which helps to build correct alignment for each position. For example, consider the pair ``A red car is next to a green house'' and ``A red car is parked near a house''. If the noun phrases in the two sentences are not correctly modeled by the contextual encoding and ``green'' is incorrectly aligned with another color word ``red'', the pair looks much less like entailment. 

In Figure \ref{fig:neutral} and Figure \ref{fig:contradiction}, we can see that lacking direct access of previous aligned features (residual vectors), especially in the final block, results in significant degradation when predicting neutral and contradiction. Previous aligned features are related to the second stage of focusing on a subset of the aligned pairs. Without correct focus, the model may ignore non-entailing pairs and attend to other trivially aligned and semantically matched pairs, which results in failure in predicting neutral and contradiction. The importance of each position can be distilled and stored in previous aligned features and helps the model to focus in latter blocks.

We can conclude from Figure \ref{fig:neutral} and Figure \ref{fig:contradiction} that when original point-wise features represented by embedding vectors are not directly accessible by alignment layers and fusion layers, the model is struggling to predict neutral and contradiction correctly. Original point-wise features connect with the final stage where semantic differences between aligned pairs are compared. Intact point-wise representations of the aligned pairs facilitate the model in the comparison of their semantic differences, which plays a vital role in predicting neutral and contradiction.

\begin{figure}
  \centering
  \begin{subfigure}{\linewidth}
    \centering
    \includegraphics[width=\textwidth]{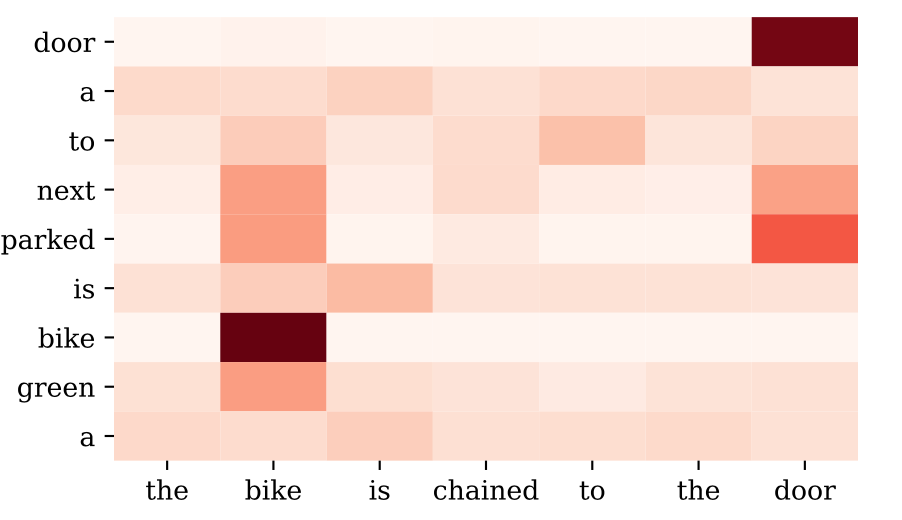}
    \caption{Alignment results in the first block}
    \label{fig:ex1}
  \end{subfigure}
  \begin{subfigure}{\linewidth}
    \centering
    \includegraphics[width=\textwidth]{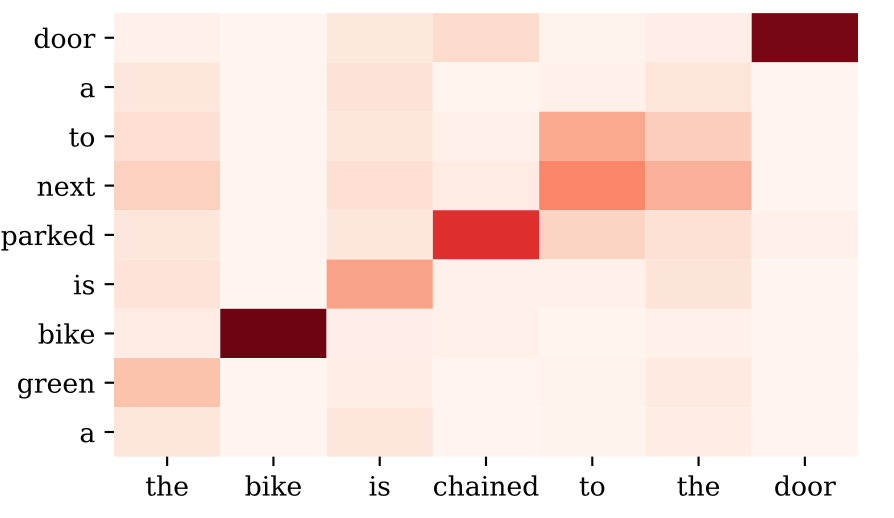}
    \caption{Alignment results in the third block}
    \label{fig:ex2}
\end{subfigure}
  \caption{A case study of the natural language inference task. The premise is ``A green bike is parked next to a door'', and the hypothesis is ``The bike is chained to the door''. }
  \label{fig:example}
\end{figure}

{\bf Case study}. We present a case study of our model to show how inter-sequence alignment results evolve in our stacked architecture. An example pair of sentences are chosen from the development set of the SNLI dataset. The premise is ``A green bike is parked next to a door'', and the hypothesis is ``The bike is chained to the door''. Figure \ref{fig:example} shows the visualization of the attention distribution (normalized $e_{ij}$ in Equation \ref{eq:aligned}) in alignment layers of the first and the last blocks. 

In the first block, the alignment results are almost word- or phrase-level. ``parked next to'' is associated mostly with ``bike'' and ``door'' since there is a weaker direct connection between ``parked'' and ``chained''. In the final block, the alignment results take consideration of the semantics and structures of the whole sentences. The word ``parked'' is strongly associated with ``chained'' and ``next to'' is aligned with ``to the'' following ``chained''. With correct alignment, the model is able to tell that although most parts in the premise entail the aligned parts in the hypothesis, ``parked'' does not entail ``chained'', so it correctly predicts that the relation between the two sentences is neutral. Our model keeps the lower-level alignment results as intermediate states and gradually refines them to higher-level ones.

\section{Related Work} \label{related}

Deep neural networks are dominant in the text matching area. 
Semantic alignment and comparison between two text sequences lie in the core of text matching. 
Early works explore encoding each sequence individually into a vector and then building a neural network classifier upon the two vectors. In this paradigm, recurrent \cite{snli}, recursive \cite{tai2015improved} and convolutional \cite{yu2014deep, tan2016improved} networks are used as the sequence encoder. The encoding of one sequence is independent of the other in these models, making the final classifier hard to model complex relations.

Later works, therefore, adopt the matching aggregation framework to match two sequences at lower levels and aggregate the results based on the attention mechanism. DecompAtt \cite{parikh2016decomposable} uses a simple form of attention for alignment and aggregate aligned representations with feed-forward networks. ESIM \cite{chen2017enhanced} uses a similar attention mechanism but employs bidirectional LSTMs as encoders and aggregators. 

Three major paradigms are adopted to further improve performance. First is to use richer syntactic or hand-designed features. HIM \cite{chen2017enhanced} uses syntactic parse trees. POS tags are found in many previous works including \citeauthor{tay2018compare} \shortcite{tay2018compare} and \citeauthor{gong2018natural} \shortcite{gong2018natural}. The exact match of lemmatized tokens is reported as a powerful binary feature in \citeauthor{gong2018natural} \shortcite{gong2018natural} and \citeauthor{kim2018semantic} \shortcite{kim2018semantic}. 
The second way is adding complexity to the alignment computation. BiMPM \cite{wang2017bilateral} utilizes an advanced multi-perspective matching operation, and MwAN \cite{tan2018multiway} applies multiple heterogeneous attention functions to compute the alignment results. 
The third way to enhance the model is building heavy post-processing layers for the alignment results. CAFE \cite{tay2018compare} extracts additional indicators from the alignment process using alignment factorization layers. DIIN \cite{gong2018natural} adopts DenseNet as a deep convolutional feature extractor to distill information from the alignment results. 

More effective models can be built if inter-sequence matching is allowed to be performed more than once. CSRAN \cite{tay2018co} performs multi-level attention refinement with dense connections among multiple levels. DRCN \cite{kim2018semantic} stacks encoding and alignment layers. It concatenates all previously aligned results and has to use an autoencoder to deal with exploding feature spaces. SAN \cite{liu2018stochastic} utilizes recurrent networks to combine multiple alignment results. This paper also proposes a deep architecture based on a new way to connect consecutive blocks named augmented residual connections, to distill previous aligned information which serves as an important feature for text matching.

\section{Conclusion}

We propose a highly efficient approach, RE2, for general purpose text matching. It achieves the performance on par with the state-of-the-art on four well-studied datasets across three different text matching tasks with only a small number of parameters and very high inference speed. It highlights three key features, namely previous aligned features, original point-wise features, and contextual features for inter-sequence alignment and simplifies most of the other components. Due to its fast speed and strong performance, the model is quite suitable for a wide range of related applications.

\bibliography{acl2019}
\bibliographystyle{acl_natbib}

\end{document}